\documentclass[11pt]{article}

\usepackage[preprint]{acl}

\usepackage{microtype}
\usepackage{booktabs}
\usepackage{graphicx}
\usepackage{xcolor}
\usepackage{xurl}

\usepackage{fontspec}
\defaultfontfeatures{Ligatures=TeX}
\newfontfamily\cyrillicfont{FreeSerif}[
  Extension = .otf,
  UprightFont = *,
  BoldFont = *Bold,
  ItalicFont = *Italic,
  BoldItalicFont = *BoldItalic,
  Script = Cyrillic
]
\newfontfamily\cyrillicfontsf{FreeSerif}[
  Extension = .otf,
  UprightFont = *,
  BoldFont = *Bold,
  ItalicFont = *Italic,
  BoldItalicFont = *BoldItalic,
  Script = Cyrillic
]
\newfontfamily\cyrillicfonttt{DejaVuSansMono}[
  Extension = .ttf,
  UprightFont = *,
  BoldFont = *-Bold,
  ItalicFont = *-Oblique,
  BoldItalicFont = *-BoldOblique,
  Scale = 0.9,
  Script = Cyrillic
]
\usepackage{polyglossia}
\setdefaultlanguage{english}
\setotherlanguage{russian}

\newif\ifshowrepolinks
\showrepolinkstrue

\newcommand{\PaperLinks}{%
  \ifshowrepolinks
    Code, data, and weights:\\
    {\footnotesize\raggedright
      \url{https://github.com/naadgob/KazNLP}\\
      \url{https://huggingface.co/datasets/naadgob/kaznlp-weights}\par}%
  \else
    Code and data will be released upon publication.%
  \fi
}

\title{Loanword or Switch? The Annotation Boundary, Not the Model,
Drives Kazakh--Russian Code-Switching Identification}

\author{
  Bogdan Savelyev \\
  Independent Researcher \\
  Kazakhstan \\
  \texttt{bogsav999@gmail.com}
}

\begin{document}
\maketitle

\begin{abstract}
Off-the-shelf LID and letter heuristics over-label Kazakh--Russian social
text as \emph{mixed}: Russian loanwords inside Kazakh look like
code-switching under a shared Cyrillic script.
We release a document-level gold LID set whose guideline
keeps integrated borrowings as Kazakh and reserves \emph{mixed} for
clause-level switches, plus a mixed-only sentiment pool used after LID
in a filter-first cascade.
On a shared LID test, FastText, Lingua, raw and windowed HeLI,
character-trigram NB, and XLM-R range from weak to strong performance.
The gap shows the bottleneck is the loanword-vs-switch annotation
boundary, not model class alone.
\PaperLinks
\end{abstract}

\section{Introduction}

Most people in Kazakhstan speak both Kazakh and Russian, and online writing
often mixes the two in one message.
We set out to build sentiment analysis for that code-switched slice, but no
open labeled set existed for the task, so we collected a corpus of over
420{,}000 Telegram, Kaspi, and 2GIS texts.
The scrape was unlabeled: to train sentiment only on genuine
\emph{mixed} messages (Kazakh and Russian used in one document), we first
needed automatic language identification (LID) at a scale that rules out
hand-filtering every row.

A first, letter-based heuristic tagged a message as \emph{mixed} whenever
it contained both Kazakh- and Russian-specific letters.
Kazakh and Russian both use Cyrillic and share most of their alphabet, and
everyday Kazakh carries many identical Russian loanwords, so almost any
Kazakh sentence looked ``mixed'' under that rule.
Off-the-shelf LID (FastText, Lingua) makes the same mistake: it cannot
tell an integrated borrowing from a real switch.

The distinction is easy to miss on a letter-level rule, but clear once
the annotation line is spelled out. Consider two short messages:

\begin{quote}
\small
\textbf{Mixed:} {\foreignlanguage{russian}{Курьер молодец, уақытында әкелді.}}\\
\hspace*{1em}(Russian clause + Kazakh clause: a genuine switch.)\\[0.35em]
\textbf{Not mixed (loanword):} {\foreignlanguage{russian}{Качествосы жақсы, арзан.}}\\
\hspace*{1em}(Kazakh morphosyntax with an integrated Russian loan,
{\foreignlanguage{russian}{качество}} + Kazakh suffix
{\foreignlanguage{russian}{-сы}}; no switch.)
\end{quote}

Without an explicit loanword-vs-switch rule, both look ``mixed'' to a
naive letter heuristic and to many off-the-shelf LID systems, because
both contain Russian and Kazakh material written in the same Cyrillic
script.

The false-\emph{mixed} problem is not only weak models; it is a failure
of the label definition.
We show this on a shared gold held-out test ($n{=}461$), comparing neural
LID with simple non-neural baselines.

Our contributions are:
\begin{enumerate}
    \item A document-level gold LID set of 3{,}076 messages (ru~/~kz~/~mixed)
    with an explicit loanword-vs-switch annotation rule.
    \item A comparison ladder on that test: FastText, Lingua, HeLI
    (raw~/~loanword-neutral~/~overlapping windows), character-trigram NB,
    and fine-tuned XLM-R; Char-3gram is near ceiling on monolingual classes
    (ru $146/150$, kz $147/150$) but recovers only $114/161$ true mixed,
    so the bottleneck is the label boundary rather than missing character
    signal.
    \item Corpus-scale application: on 331{,}468 documents the filter keeps
    $4.9\%$ as mixed, against a keyword heuristic with
    ${\approx}1.66\%$ precision on its Telegram mixed pool.
    \item A filter-first sentiment head (pos/neg) trained on a mixed-only
    gold pool and run only after LID; reported as downstream utility, not
    as the main claim.
\end{enumerate}

\section{Related Work}

The theoretical line behind our label rule is the long-standing distinction
between borrowing and code-switching \citep{poplack1980}: integrated
loanwords belong to the recipient language, while productive switches do
not. Computational work often collapses that line. Shared or ambiguous
lexicon is treated as a special tag in some Arabic CS corpora
\citep{wintner2023shared}; we keep a document-level three-way label
(\emph{ru}/\emph{kz}/\emph{mixed}) and push the same ambiguity into the
annotation guideline instead.

Language identification is surveyed extensively by
\citet{jauhiainen2019survey}. For short, noisy, or multilingual spans,
word-list systems such as HeLI and language-set identification over
overlapping windows \citep{jauhiainen2015} remain strong, interpretable
baselines. Large-coverage neural or n-gram LID tools (OpenLID, GlotLID)
work well on clean monolingual text but are known to overestimate accuracy
on web data \citep{burchell2023openlid,kargaran2023glotlid}. Our setting
adds a harder failure mode: Kazakh and Russian share Cyrillic, so
loanword-bearing Kazakh looks mixed to letter heuristics and to many
off-the-shelf detectors.

Character n-grams have a long history for Turkic and web-scale LID
\citep{baisa}; we include a smoothed character-trigram Naive Bayes
baseline for that reason. Closely related contact pairs with the same
boundary problem appear in Kyrgyz--Russian work and in Kyrgyz web corpora
contaminated by neighboring Turkic text; we treat those as parallel
motivation rather than as datasets we reuse.

Code-switching workshops (CALCS) and variation shared tasks (VarDial) are
natural homes for underrepresented pairs and for similar-language
confusability. We do not frame kk--ru as dialect identification: the two
are distinct languages in contact. The scientific claim is narrower---that
an explicit loanword-vs-switch annotation rule changes what automatic LID
can honestly measure on a shared-script pair.

\section{The Gold Set}

Table~\ref{tab:gold-sets} summarizes the two labeled resources.
LID is document-level ru/kz/mixed; SA is pos/neg on mixed reviews only.
Both were annotated solo (no IAA).

\begin{table}[t]
\centering
\setlength{\tabcolsep}{3.5pt}
\small
\begin{tabular}{@{}lrrp{2.55cm}@{}}
\toprule
\textbf{Set} & \textbf{$n$} & \textbf{Gold} & \textbf{Rule} \\
\midrule
LID
  & 3076 & 3076
  & borrow${\to}$kz; switch${\to}$mixed \\
SA
  & 4411 & 3529
  & pos/neg; mixed only \\
\bottomrule
\end{tabular}
\caption{Gold resources. LID classes: mixed 1077~/~ru 1000~/~kz 999
(split 2691/462/461). SA ${\sim}$50/50 pos/neg (split 3334/526/525;
882 synthetic). Solo annotation; no IAA.}
\label{tab:gold-sets}
\end{table}

\section{Experimental Setup}

All LID systems are evaluated on the same held-out gold test
($n{=}461$; train $2{,}691$~/~val $462$~/~test $461$).
We report accuracy, macro-F1, and mixed-class precision/recall.
Supervised models (FastText, Char-3gram NB, XLM-R LID) train on the
gold training split only.

\paragraph{HeLI ladder.}
As an interpretable non-neural baseline we use HeLI through
\emph{heliport}.\footnote{Rust/Python port of HeLI-OTS:
\url{https://github.com/ZJaume/heliport}}
The ladder below follows advice from Tommi Jauhiainen in email
correspondence, adapted here to short kk--ru social texts and our
document-level \emph{mixed} label.

The first step is a \textbf{loanword-neutral re-identification}
pass: build a list of 100\%-identical Russian borrowings used in
Kazakh, remove (or score neutral) those tokens, and re-run HeLI
(\textbf{HeLI+neutral}).
That cleans shared vocabulary but does not by itself detect genuine
code-switching.
On our test that prediction held: HeLI raw reaches $0.697$ macro-F1,
while HeLI+neutral is slightly worse ($0.683$), leaving $80$
gold-\emph{mixed} documents still tagged \emph{rus} after stripping.

To recover those switches, we then adapt HeLI's language-set idea
\citep{jauhiainen2015} down from long documents to short UGC lines:
after the loanword strip, split each text into overlapping two-,
three-, or longer word windows, run heliport on each window, and
output \emph{mixed} if more than one language appears (optionally with
a higher count threshold).
A small \textbf{grid search} over window sizes and
$\mathrm{min\_count}$ tracks the typical length of switches in this
data.
We implement that as \textbf{HeLI+windows}; the best setting on our
grid is sizes $(2,3)$ with $\mathrm{min\_count}{=}1$, which raises
macro-F1 from $0.697$ (raw) to $0.869$ and flips $69/80$ of the
residual mixed-as-\emph{rus} bucket to \emph{mixed}.

\paragraph{Other systems.}
The character baseline is Multinomial Naive Bayes over raw-text
character trigrams
(\texttt{CountVectorizer(analyzer='char', ngram\_range=(3,3))},
Laplace $\alpha{=}1$).
Off-the-shelf comparators are FastText and Lingua (v1/v2).
The neural LID model is XLM-RoBERTa fine-tuned on Gold LID
(XLM-R LID v1/v2; v2 is the main neural result).

Sentiment analysis (SA) is a separate binary head (pos/neg) on the
filtered \emph{mixed} review slice only (filter-first cascade),
evaluated on its own hold-out ($n{=}525$), not on the LID test.

\section{Results: The Ladder}

Table~\ref{tab:full-ladder} reports the full comparison.

\begin{table*}[t]
\centering
\setlength{\tabcolsep}{6pt}
\small
\begin{tabular}{@{}lcccc@{}}
\toprule
\textbf{Model} & \textbf{Acc} & \textbf{Macro-F1} & \textbf{P} & \textbf{R} \\
\midrule
\multicolumn{5}{@{}l}{\textit{LID (ru~/~kz~/~mixed; $n{=}461$). P/R $=$ mixed.}} \\
\midrule
FastText v1              & 0.649 & 0.632 & 0.557 & 0.335 \\
FastText v2              & 0.716 & 0.709 & 0.658 & 0.491 \\
HeLI raw                 & 0.703 & 0.697 & 0.589 & 0.534 \\
HeLI+neutral             & 0.690 & 0.683 & 0.573 & 0.491 \\
HeLI+windows (2+3, min1) & 0.872 & 0.869 & 0.925 & 0.689 \\
Char-3gram NB            & 0.883 & 0.880 & 0.942 & 0.708 \\
Lingua v1                & 0.846 & 0.850 & 0.739 & 0.863 \\
Lingua v2                & 0.889 & 0.886 & 0.768 & 0.988 \\
XLM-R LID v1             & 0.959 & 0.959 & 0.938 & 0.944 \\
XLM-R LID v2             & \textbf{0.965} & \textbf{0.966} & 0.950 & 0.950 \\
\midrule
\multicolumn{5}{@{}l}{\textit{SA (pos~/~neg on mixed; $n{=}525$). P/R $=$ positive.}} \\
\midrule
XLM-R SA v1 (best)       & \textbf{0.973} & \textbf{0.973} & 0.977 & 0.969 \\
XLM-R SA v2              & 0.962 & 0.962 & 0.962 & 0.962 \\
\bottomrule
\end{tabular}
\caption{Full model ladder. Top: document-level LID. Bottom: SA on the
filtered code-switched slice only (filter-first cascade).
For SA v1, negative-class P/R ${=}0.970/0.977$; for SA v2 both ${=}0.962$.
Most SA labels are LLM-drafted.}
\label{tab:full-ladder}
\end{table*}

Several patterns stand out from Table~\ref{tab:full-ladder}.
First, stripping shared loanwords alone does not help HeLI:
HeLI+neutral (macro-F1 $0.683$) is slightly worse than HeLI raw ($0.697$).
Lexical neutralization without structure is not enough for document-level
switch detection.
Second, overlapping windows carry most of the non-neural gain.
HeLI+windows (sizes $2{+}3$, $\mathrm{min\_count}{=}1$) jumps to $0.869$
macro-F1 and mixed precision $0.925$; of the $80$ gold-mixed documents
that stayed \emph{rus} after the strip, $69$ flip to \emph{mixed}.
Third, the character-trigram NB is the strongest simple orthographic
baseline ($0.880$ macro-F1). Monolingual recall is near ceiling
(ru $146/150$, kz $147/150$), yet mixed recall is only $0.708$
($47$ misses: $20{\to}\mathrm{ru}$, $27{\to}\mathrm{kz}$).
The bottleneck is the loanword-vs-switch boundary, not missing
character signal.
Fourth, off-the-shelf LID shows a different trade-off: Lingua v2 reaches
mixed recall $0.988$ but precision only $0.768$, so it over-tags mixed
and is a weak corpus filter. FastText v2 improves on v1 but stays near
HeLI raw on macro-F1 (${\sim}0.71$).
Fifth, XLM-R LID v2 is the only system with balanced mixed P/R near
$0.95$ (macro-F1 $0.966$). The gap over Char-3gram is contextual
modeling of the whole message, not merely ``neural vs non-neural.''
Sixth, sentiment analysis is a downstream head on the filtered mixed
slice. XLM-R SA v1 is best (accuracy~/~macro-F1 $0.973$ on $n{=}525$);
v2 is close but lower ($0.962$). These numbers measure cascade utility
after LID, not a standalone sentiment benchmark, and most tone labels
are LLM-drafted.

\section{Error Analysis}

Of the 80 gold-\emph{mixed} documents that HeLI still tagged \emph{rus}
after loanword stripping, HeLI+windows (sizes $2{+}3$,
$\mathrm{min\_count}{=}1$) flips 69 to \emph{mixed}.
We inspected the remaining 11 by hand: tokens after strip, every
window code from heliport, and isolated probes of the Kazakh spans.

All 11 fail the same way. After stripping, no 2- or 3-word window
ever receives code \emph{kaz}, so the language-set vote never sees a
Kazakh side and cannot emit \emph{mixed}.
None of the cases fail because the document is too short, and none
fail because a \emph{kaz} window existed but lost the vote.
The limit sits in heliport on short windows where Kazakh is glued to
Russian neighbors or written with informal spelling.

\paragraph{Single-word Kazakh insertions (5/11).}
One Kazakh content token sits inside an otherwise Russian matrix
({\foreignlanguage{russian}{өсек}},
{\foreignlanguage{russian}{қоқыс}},
{\foreignlanguage{russian}{сәлем}},
{\foreignlanguage{russian}{үйренеді}},
{\foreignlanguage{russian}{бұйырса}}).
With minimum window size 2, that token always shares a window with
one or two Russian neighbors, and heliport returns \emph{rus} for
spans such as
``{\foreignlanguage{russian}{здесь үйренеді}}'' or
``{\foreignlanguage{russian}{это өсек}}''.
A pure Kazakh window never forms.
Probed alone at size 1, the same five tokens all map to \emph{kaz}:
the material is recognizable; windowing at ${\ge}2$ is what buries it.

\paragraph{Short Kazakh spans misread as other languages (4/11).}
Two-word Kazakh chunks get a related Turkic or Slavic code, which our
vote discards because we only keep \emph{kaz}/\emph{rus}:
{\foreignlanguage{russian}{терең ой}}${\to}$\emph{kir},
{\foreignlanguage{russian}{біткенше күлдім}}${\to}$\emph{ukr}
(standard spelling; the 200-language identifier confuses short Kazakh
with Kyrgyz or Ukrainian), and
{\foreignlanguage{russian}{қудай сактады}}/
{\foreignlanguage{russian}{кудай сақтасын}}${\to}$\emph{kir}, where
informal spelling drops
{\foreignlanguage{russian}{қ}}/
{\foreignlanguage{russian}{ұ}}
and removes the main Kazakh cue.
Rewritten as
{\foreignlanguage{russian}{құдай сақтасын}} or
{\foreignlanguage{russian}{құдай сақтады}},
heliport returns \emph{kaz}.

\paragraph{One token heliport never tags as Kazakh (1/11).}
{\foreignlanguage{russian}{шапшаң}} alone scores as \emph{mhr}
(Meadow Mari).
Size-1 windows would not recover it; the miss is in the identifier,
not in the window schedule.

\paragraph{Likely label noise (1/11).}
{\foreignlanguage{russian}{«Порог растет, а зп нет»}} has no Kazakh
letters and no Kazakh words ($kz\_signal{=}\mathrm{False}$), yet gold
marks it \emph{mixed}.
We treat it as a candidate annotation error pending a second look at
the source thread.

\paragraph{Size-1 windows as a diagnostic.}
Adding size-1 windows (sizes $(1,2,3)$) raises macro-F1 from $0.869$
to $0.885$ and mixed recall from $0.689$ to $0.776$, but mixed
precision falls from $0.925$ to $0.880$
($kz{\to}mixed$ errors $8{\to}14$, $ru{\to}mixed$ $1{\to}3$).
Single-token votes catch insertional switches and also punish clean
Kazakh that happens to contain one Russian-looking token.
The advice that motivated the ladder was ``two word, three word or
longer''; size 1 is outside that definition and closer to per-token LID.
We therefore keep $(2,3)$ as the HeLI+windows rung on the ladder and
report size 1 only here, as an explicit trade-off rather than as a
replacement.

\paragraph{Takeaway.}
Windowed HeLI recovers clause-level switches; by construction it
misses one-word insertions when the minimum window is 2.
At least four of the eleven residuals are not about windows at all:
heliport confuses short Kazakh with Kyrgyz (or Mari), and informal
spelling without
{\foreignlanguage{russian}{қ}}/
{\foreignlanguage{russian}{ұ}}
makes that worse.
Char-3gram fails on a related boundary: of 161 true mixed documents
it reads 47 as monolingual ($20{\to}\mathrm{ru}$, $27{\to}\mathrm{kz}$),
again short insertions and loanword-bearing spans.
Both non-neural baselines stop at the same loanword-vs-switch line;
XLM-R LID crosses it with full-message context.

\section{Corpus-Scale Application}

A gold test of $n{=}461$ only shows that the filter works on held-out
labels. The practical question is what happens when the same XLM-R LID
v2 model scores a full scrape.

We build a pooled corpus of Telegram comments, Kaspi reviews, and 2GIS
reviews across Kazakhstani cities, then deduplicate and clean it into
\texttt{main.csv}: $331{,}468$ documents.
XLM-R LID v2 assigns
$281{,}409$ \emph{ru}, $33{,}695$ \emph{kz}, and $16{,}364$ \emph{mixed}.
That is a mixed rate of $4.9\%$.

The contrast with the early keyword heuristic is the whole point of the
resource.
On the Telegram slice alone ($241{,}576$ messages), a narrow
``Kazakh-letter + Russian-letter'' rule flagged $27{,}628$ rows as
\emph{mixed}; hand-checking that pool kept only $460$ genuine switches
(precision ${\approx}1.66\%$).
A later FastText-v2 pass over a larger scrape still sat near $2\%$
heuristic precision on its mixed predictions.
Once the gold-trained filter replaces those rules, the corpus-level
mixed share collapses from ``almost everything bilingual'' to roughly
one document in twenty.

Two caveats keep the claim honest.
First, the $16{,}364$ mixed labels are model predictions, not a second
full hand audit; gold mixed precision/recall near $0.95$ on $n{=}461$
does not automatically transfer to every corpus row.
Second, the $27{,}628/460$ audit covers the Telegram heuristic pool,
not the entire $331$k file.
Even with those limits, the order of magnitude is stable enough to
change what one can claim about code-switching prevalence in this
domain: loanword-heavy Kazakh was inflating the mixed count, and a
document-level loanword-vs-switch rule cuts that inflation down.

\section{Conclusion}

False \emph{mixed} on Kazakh--Russian social text is mostly a labeling
problem, not a model problem.
Once the guideline treats integrated Russian borrowings as Kazakh and
reserves \emph{mixed} for clause-level switches, the same held-out gold
test ($n{=}461$) separates the systems cleanly: FastText and raw HeLI sit
near $0.70$ macro-F1, HeLI+windows (language-set adaptation with
overlapping sizes $2{+}3$) reaches $0.869$, Char-3gram NB $0.880$, and
XLM-R LID v2 $0.966$.
Char-trigram almost never confuses monolingual ru/kz, yet recovers only
$70.8\%$ of true mixed; the bottleneck is the loanword-vs-switch line,
which full-message context closes.

At corpus scale the same filter changes the story you can tell about
the data.
On $331{,}468$ documents, XLM-R LID v2 keeps $16{,}364$ ($4.9\%$) as
\emph{mixed}, against a Telegram keyword heuristic that flagged
$27{,}628$ rows with only $460$ hand-confirmed switches
(${\approx}1.66\%$ precision).

We release the gold LID and SA pools, the HeLI ladder code, and the
trained weights.
Next steps are a second annotator for IAA, a morphology cue
(Apertium) for bare Russian nouns inside Kazakh, and polarity models
that run only on the filtered mixed slice rather than on
loanword-inflated ``mixed'' noise.

\section*{Limitations}

Annotation is solo; we report no inter-annotator agreement.
That is the main weakness of the resource, and several external
readers already flagged it.

The $16{,}364$ corpus mixed labels are XLM-R predictions, not a full
hand audit.
Gold mixed P/R near $0.95$ on $n{=}461$ should not be read as a
guarantee on every corpus row.

Sentiment labels are mostly LLM-drafted then lightly audited.
We therefore treat SA metrics as cascade utility after LID, not as a
primary claim about human sentiment annotation.

Labels are document-level.
That flattens token-level gray zones (bare Russian nouns with no
Kazakh morphology; short insertional switches that windowed HeLI
misses by construction).
Informal spelling without
{\foreignlanguage{russian}{қ}}/
{\foreignlanguage{russian}{ұ}}
also pushes short Kazakh spans into neighboring Turkic codes in
heliport.

\ifshowrepolinks
\section*{Acknowledgments}
This work began as an individual capstone project at Samsung Innovation Campus.
I thank Tommi Jauhiainen for detailed advice on the HeLI ladder
(loanword-neutral re-identification, overlapping windows after the strip,
and grid search over window size and $\mathrm{min\_count}$), and for
pointing me to heliport and the language-set identification line of work.
I also thank Anton Alekseev, Jonathan Washington, Jonathan Dunn, Maite Heredia,
and Nikola Ljube{\v{s}}i{\'c} for framing advice and for pushing the
character-trigram baseline.
\fi

\bibliography{custom}

\end{document}